\documentclass[manuscript]{acmart}
\AtBeginDocument{%
  \providecommand\BibTeX{{%
    \normalfont B\kern-0.5em{\scshape i\kern-0.25em b}\kern-0.8em\TeX}}}

\usepackage{booktabs}

\usepackage{subcaption}
\graphicspath{ {./figures/} }

\begin{document}

\title{Identifying Introductions in Podcast Episodes from Automatically Generated Transcripts}

\author{Elise Jing}
\email{yjing@pandora.com}

\author{Kristiana Schneck}
\email{kschneck@pandora.com}

\author{Dennis Egan}
\email{Dennis.Egan@siriusxm.com}

\author{Scott A. Waterman}
\email{swaterman@pandora.com}
\affiliation{%
  \institution{Sirius XM}
  \streetaddress{1221 Avenue of the Americas 37th Floor}
  \city{New York}
  \state{NY}
  \country{USA}
  \postcode{10020}
}

\renewcommand{\shortauthors}{Jing, et al.}

\begin{abstract}
As the volume of long-form spoken-word content such as podcasts explodes, many platforms desire to present short, meaningful, and logically coherent segments extracted from the full content.
Such segments can be consumed by users to sample content before diving in, as well as used by the platform to promote and recommend content.
However, little published work is focused on the segmentation of spoken-word content, where the errors (noise) in transcripts generated by automatic speech recognition (ASR) services poses many challenges. 
Here we build a novel dataset of complete transcriptions of over 400 podcast episodes, in which we label the position of introductions in each episode. 
These introductions contain information about the episodes' topics, hosts, and guests, providing a valuable summary of the episode content, as it is created by the authors.
We further augment our dataset with word substitutions to increase the amount of available training data. 
We train three Transformer models based on the pre-trained BERT~\shortcite{devlin2018bert} and different augmentation strategies, which achieve significantly better performance compared with a static embedding model, showing that it is possible to capture generalized, larger-scale structural information from noisy, loosely-organized speech data. 
This is further demonstrated through an analysis of the models' inner architecture. 
Our methods and dataset can be used to facilitate future work on the structure-based segmentation of spoken-word content.
\end{abstract}

\maketitle

\section{Introduction} \label{sec:introduction}

As digital spoken-word content, such as podcasts, increases in quantity and availability, content creation and distribution platforms face many challenges related to organizing, understanding, and promoting their content. 
The large scale of spoken-word content also presents challenges to users who would like to explore the catalogs and choose content to listen to.
Many podcasts are long and require focused attention, so users may spend longer in deciding whether to commit to them. 
Video platforms, including Netflix and YouTube, address this issue by editing and producing previews and shorter clips from long-form content. 
Such previews and clips typically consist of short segments that are meaningful, entertaining, logically coherent, and self-contained, providing a compelling way for users to sample content before committing to watching the full episode or show. Additionally, they can be used to promote content off-platform, and further increase users' engagement with the content by personalizing recommendations and user interfaces~\cite{covington2016youtube,amat2018netflix}. Automatic methods for video segmentation (e.g. ~\cite{rotman2018video}) are commonly used in editing and content creation tools, and are even offered as subscription APIs~\cite{rekognition}. However, similar segmentation tools for speech content are currently not available. Methods to automatically identify reusable segments from longer speech content have clear practical applications in summarization, trailers, and highlights and quote extraction.


In this work, we focus on the problem of identifying the introductory part of a podcast episode based on automatic speech recognition (ASR) transcriptions. 
An introduction in a podcast episode typically describe the episode's main subject(s), contents, and speakers, and is a good representation of the whole episode. Dialectically, the introduction is used to inform and prepare the listener for the performance ahead.
Listening to a podcast episode's introduction can often stimulate listeners' interest and help them decide whether to listen to the whole episode. Additionally, the introduction can be used in recommendation systems or off-platform promotion to attract users to the content. 

From a linguistic perspective, the task of identifying introductions is different from many common segment extraction tasks in that an introduction is recognizable based on its structural uniqueness within the longer content, rather than topical differences. 
While an introduction shares similar vocabulary and topics with the rest of the episode, a human listener can usually recognize the macro-level discourse structure that defines the introduction. 
In the computational studies of discourse structure, most work is focused on detailed phrasal elements and detection of elementary discourse units (EDUs), with application on specific tasks such as argument detection~\cite{mochales2011argumentation} and dialogue generation~\cite{hovy1993automated}. 
Although recurrent neural network (RNN) and attention models have recently been applied to build discourse-aware models for segmentation and summarization tasks~\cite{wangetal2018toward, cohan2018discourse, xu2019discourse}, they also focus on EDUs or small, clausal units, rather than larger-scale units similar to what we tackle here.

Mining spoken-word data in the form of ASR transcripts poses many additional challenges. 
Most natural language processing (NLP) techniques assume well-structured, written text divided into sentences, while ASR transcripts explicitly represent speakers' disfluencies, restarts and other errors. Neither do they have reliable punctuation or paragraphing. ASR transcripts are also affected by word errors caused by out-of-vocabulary terms or failed recognition. Moreover, difficulty in speaker detection makes it hard to identify the conversational exchanges. While existing work have addressed the segmentation of ASR data based on topic changes~\cite{bouchekif2015diachronic,chifu2016segchain} and dialogue exchanges~\cite{purver2006unsupervised,hsueh2010combining}, little work has focused on the structural segmentation of ASR data necessary for identifying these introductory segments.


While many corpora have been created for topic-based text segmentation~\cite{pak2018text}, the relatively few corpora for text segmentation based on discourse structure focus on sentence-level connectives~\cite{webber2019penn} or EDUs under the Rhetorical Structure Theory (RST) framework~\cite{carlson2001discourse}, and no dataset is available for the structure-based segmentation of ASR transcripts to the best of our knowledge. We therefore create a new dataset of labeled podcast transcripts for our task. After obtaining transcripts using ASR tools, we recruited lightly-trained volunteers to annotate the introductions, obtaining labeled data for 417 podcast episodes. We explore the annotator agreement, showing that human annotators achieve reasonable agreement on the locations and components of introductions. For the details, see Section~\ref{sec:data}\footnote{The dataset and code that we create will be made public.}.

Inspired by recent works on text segmentation using neural network models~\cite{badjatiya2018attention,li2018segbot,salloum2017automated}, we formulate our task as a supervised sequence labeling task. 
We train our models to label each token in the text, and find the best split position based on the token labels, using fine-tuning over a pre-trained BERT model~\cite{devlin2018bert}.
To highlight that our models recognize the structures of discourse data, rather than relying only on lexical cues, we compare to a baseline created using GloVe embeddings~\cite{pennington2014glove}. 
We also apply two data augmentation strategies to increase the amount of available training data (see Section~\ref{sec:approach} for details). 
Our models are evaluated using two metrics: the accuracy of identifying the segmentation boundaries (\emph{accuracy}), and the overlap between the predicted introduction and the gold standard (\emph{overlap score}).  

We find that our models outperform the baseline by significant margins. Compared to the base BERT model, data augmentation improves the accuracy by up to three percentage points, as well as decreasing the variance in model performance.
Moreover, while the baseline model performs poorly on data that is structurally different from the training data, our models show an ability to generalize (see Section~\ref{sec:results}). 
We further analyze the learned hidden representations within our models, demonstrating that they are able to learn structural information in additional to topical or lexical cues. 
Our methods and dataset can be used to facilitate future work in this domain.

\section{Data} \label{sec:data}

Our dataset consists of podcast transcripts collected using Google's speech-to-text service. 
Our work focuses on English content, and we use the transcriptions produced by the ASR systems without post-editing or clean-up. 
In order to create a varied dataset, we collect recent episodes from popular programs across 20 topical categories on our platform for manual labeling. 
At the time of submission, 417 episodes have been annotated.

A group of annotators were recruited to label the dataset. Each annotator listens to a podcast episode while looking at its transcript. 
Annotators are instructed to identify the \emph{episode introduction}, which is a short description of a specific episode's topic, host, guest, or other important subjects discussed in it. 
Podcasts may also contain \emph{program introductions} which give on overview of the program as a whole. 
These are often trivial to identify, as they repeat from episode to episode.
We focus on the episode introductions\footnote{Except in this paragraph, we use the word ``introductions" to indicate episode introductions in this paper.}. 

We ask the annotators to label the starting and ending words of each of the episode introductions, or mark ``none" if they are not present. We do not provide a detailed guideline of what is required for an introduction, but encourage the annotators to use their own judgement in order to obtain more spontaneous reaction to the data.

\begin{figure}
  \centering
    \includegraphics[width=0.4\textwidth]{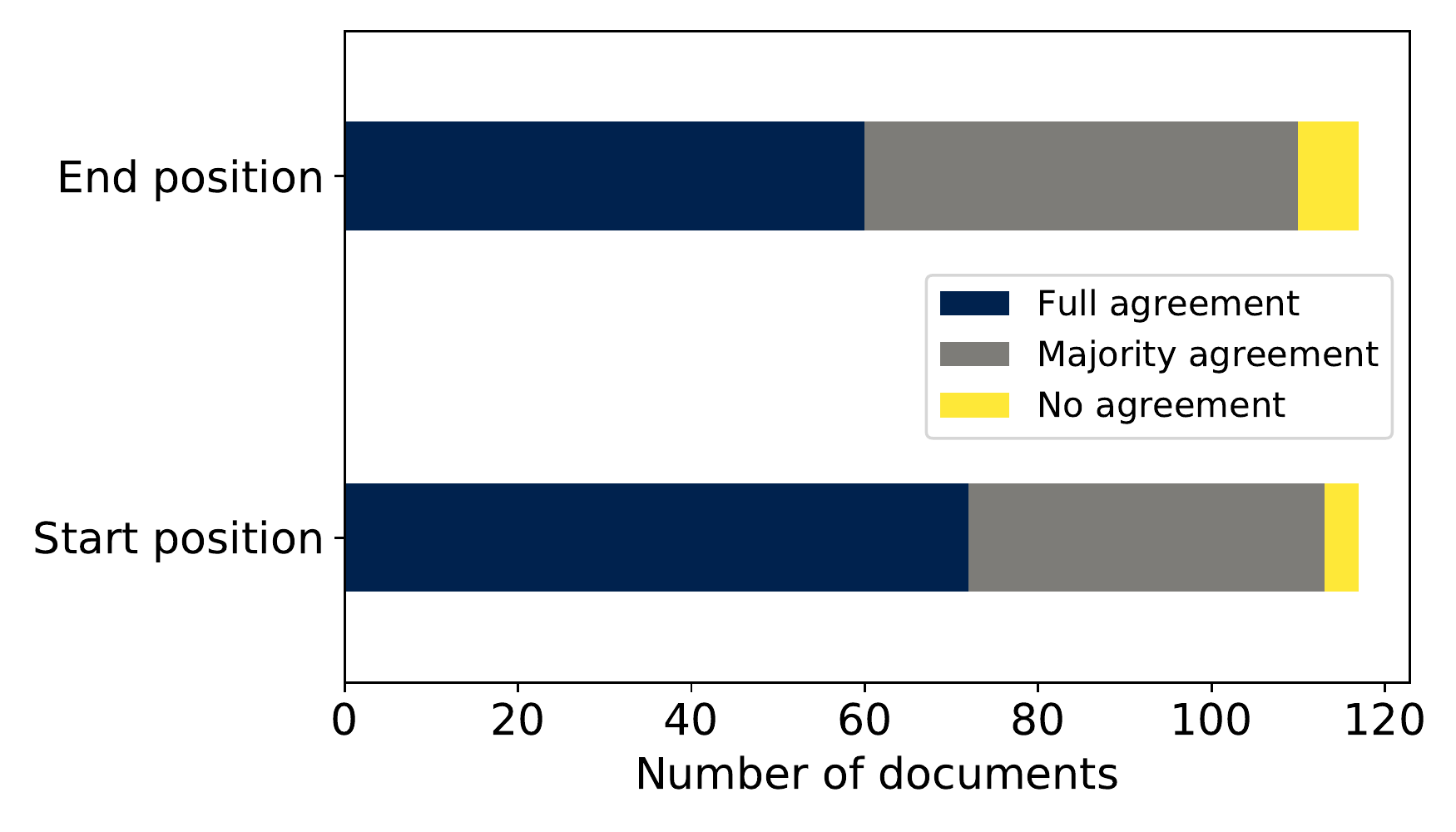}
      \caption{Agreement on podcast episode introductions with annotation from three annotators. The majority of episodes have a perfect or majority agreement, and few have no agreement.}
    \label{fig:annotation_agreement}
\end{figure}

\begin{figure*}
  \centering
  \vspace{-0.4cm}
    \includegraphics[width=0.3\textwidth]{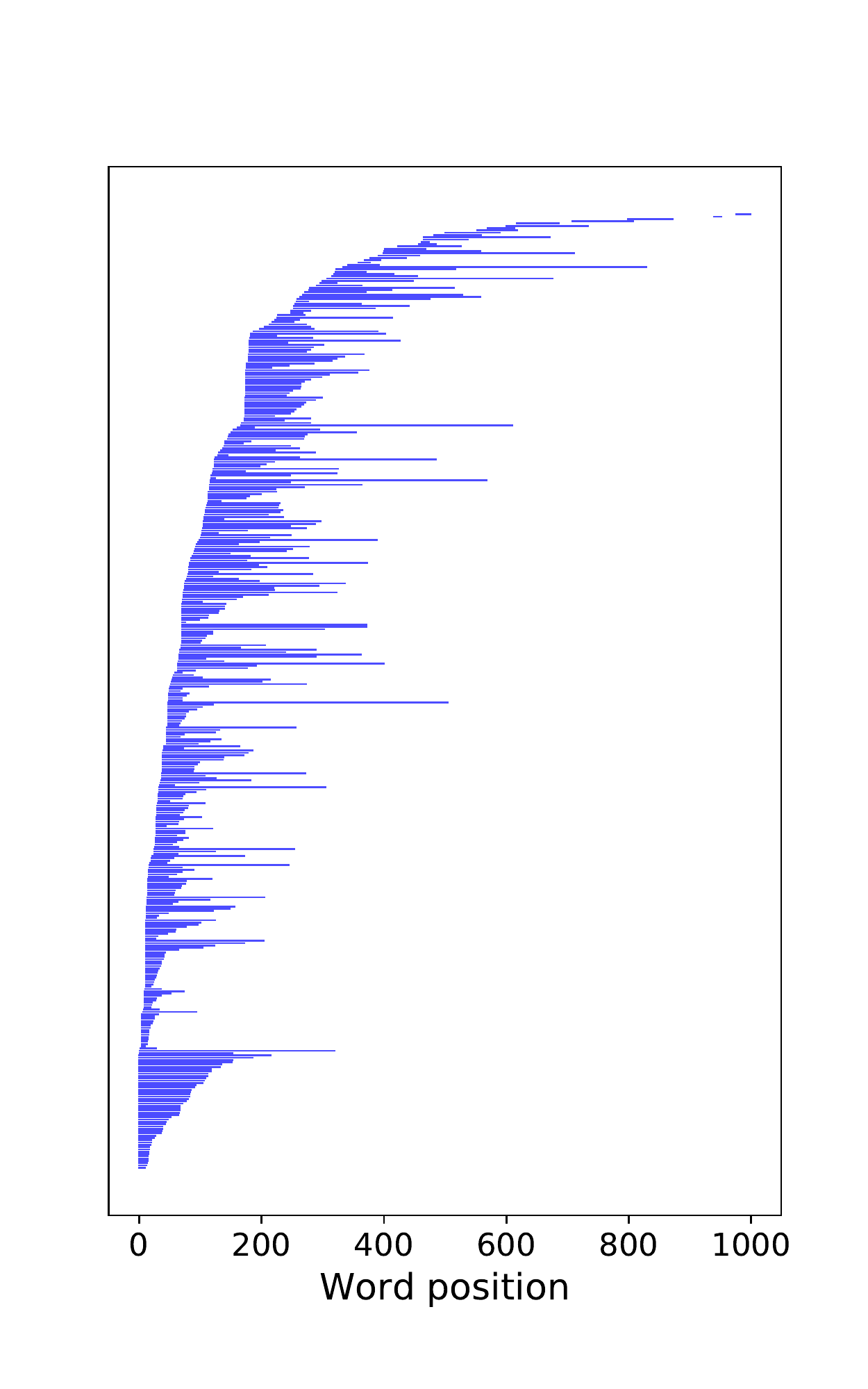}
    \vspace{-0.3cm}
      \caption{Locations of episode introductions in the transcripts. Each line shows the start, end, and duration of an episode's introduction by word positions, sorted by start position}
    \label{fig:start_end_position}
\end{figure*}

Since ASR can be prone to errors, an annotator may have trouble identifying or labeling the exact start and end of an introduction. Even if the transcription is perfect, annotators may disagree on what constitutes an introduction. Taking account of these issues, we have a number of episodes labeled by three independent annotators.

We examine the annotator agreement on 117 episodes with three annotations. We define the gold standard to be majority agreement---where at least two out of three annotators agree. If all three annotators agree, we consider it a perfect agreement. Because of the noise in the data, if two labels differ by less than 2 seconds, we consider them as having an agreement. Figure \ref{fig:annotation_agreement} shows the number of perfect and majority agreements for these episodes. 
For the annotations labeling the start of introductions, 72 out of the 117 episodes have perfect agreement. Forty-one episodes have majority agreement, and only 4 have no agreement. If we only consider episodes with perfect or majority agreement, we obtain 113 episodes or 96.6\% of the data. Similarly, considering the agreement on the end position, we are left with 110 episodes or 94.0\% of the data. For the podcast programs with 100\% annotator agreement, we include all episodes in these programs in our dataset even if they have not been labeled by three independent annotators.

From the annotated data, we notice that the structures of podcast episodes are not consistent. For example, some episodes have program introductions before episode introductions, and some vice versa. A number of episodes have no introduction at all. Music and advertisements may also appear before or after the introductions. Figure~\ref{fig:start_end_position} shows the locations of episode introductions. We found these locations to vary widely, with episode introductions starting or ending as late as near 1,000 words into the transcripts.

While the formats of podcasts vary, the episodes in a single podcast program usually share similar topic(s) and format(s). For example, all episodes in the \emph{B\&H Photography Podcast}\footnote{\url{https://www.bhphotovideo.com/explora/podcasts}}
are about photography. All episodes in the \emph{Song Exploder}\footnote{\url{http://songexploder.net/}}
program share a similar structure: introduction, a song, and then an interview with the creator(s) of the song. Intuitively, it will be much easier for a model to perform well on given episodes if it is trained on other episodes in the same program. We therefore stratify the data, leaving 5\% of the programs out and use all episodes in these programs as ``test set of unseen programs", and another 5\% of all programs as validation set. From the rest of programs, we further keep 10\% of their episodes as ``test set of seen programs" and another 10\% as validation set. The rest of the dataset is used as training set. The number of episodes in the training and test sets are summarized in Table \ref{table:data_stats}.

\begin{table*}[h]
    \centering
    \begin{tabular}{llcc}
    \hline &  \textbf{Training set size} & \textbf{Test set size} & \textbf{Validation set size}\\ \hline
    \# of episodes & 315 & 39 (seen programs)/28 (unseen programs) & 35 (seen programs)/28 (unseen programs)\\
    \# of tokens & 6,129,679   & 797,227  (seen)/184,675  (unseen) & 613, 739 (seen) / 184, 675 (unseen)\\

    \hline
    \end{tabular}
    \caption{ \label{table:data_stats} Number of episodes and tokens in the training and test sets.}
\end{table*}

\section{Approach}
\label{sec:approach}

\begin{figure*}
    \centering
    \begin{subfigure}[h]{0.8\textwidth}
        \includegraphics[width=\textwidth]{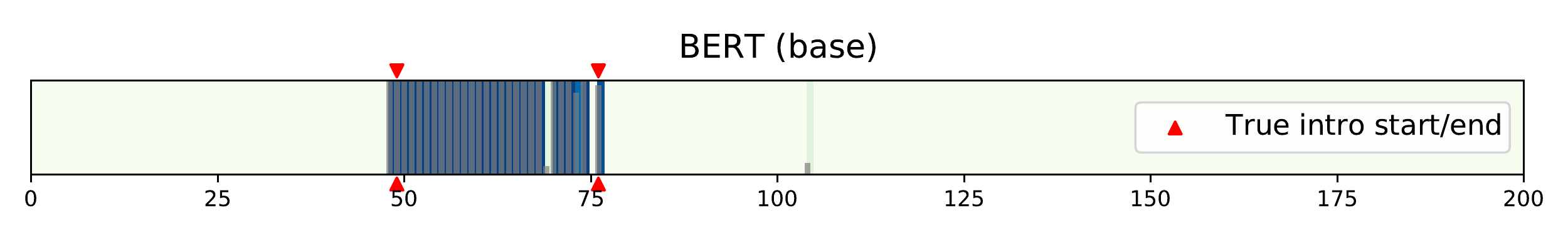}
        \vspace{-0.3in}
        \caption{}
        \label{fig:score_dist_bert_1}
    \end{subfigure}
    \begin{subfigure}[h]{0.8\textwidth}
\includegraphics[width=\textwidth]{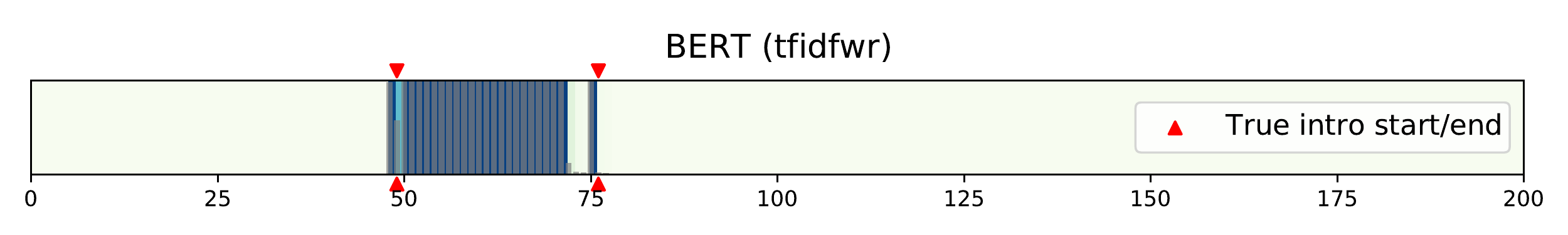}
        \vspace{-0.3in}
        \caption{}
        \label{fig:score_dist_bert_2}
    \end{subfigure}
    \begin{subfigure}[h]{0.8\textwidth}
    \includegraphics[width=\textwidth]{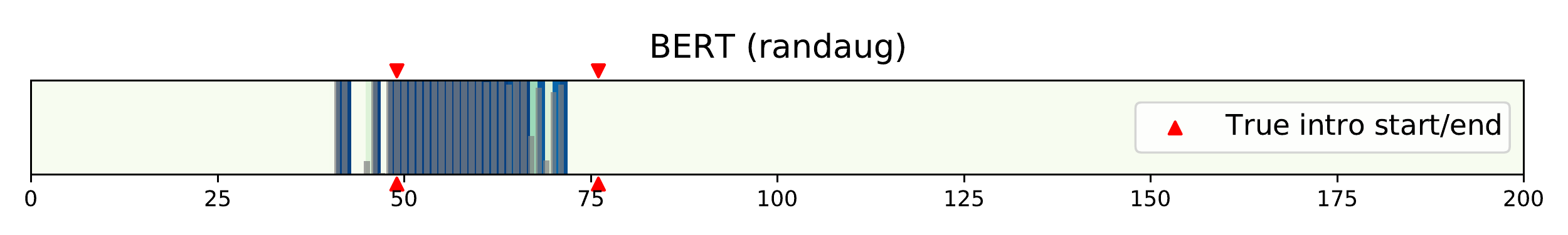}
        \vspace{-0.3in}
        \caption{}
        \label{fig:score_dist_bert_3}
    \end{subfigure}
    \begin{subfigure}[b]{0.8\textwidth}
        \includegraphics[width=\textwidth]{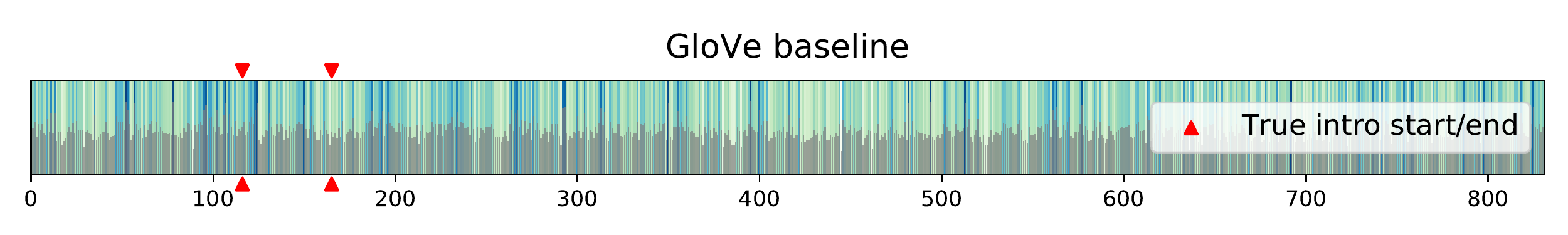}
        \vspace{-0.3in}
        \caption{}
        \label{fig:score_dist_baseline}
    \end{subfigure}
    \caption{Distribution of the models' prediction scores. The x-axis shows the index of tokens. Spans are truncated to show details of the introductions better. Each vertical line is the probability for being \texttt{Is-intro} for a token. The darker lines indicate higher probabilities, while the gray bars show the raw probability scores. The ground truth boundaries are indicated by triangles.}\label{fig:score_dist}
\end{figure*}

We start with learning contextualized vector representations for each token in the transcript by fine-tuning a BERT model. 
We first tokenize the documents with BERT's WordPiece tokenizer. 
If a document is longer than 512 tokens, we divide it into overlapping spans using a sequence length of 512 with 128 overlapping tokens between spans following the practice in~\citet{devlin2018bert}. The spans are re-merged after training using a maximum minimum method described in the same paper. We then train a fully connected layer to assign the probability of each token belonging to one of two classes \texttt{Is-intro} or \texttt{Not-intro}. As the predicted probability score falls between $0$ and $1$, the tokens predicted with high \texttt{Is-intro} probabilities may be found throughout a document (see Figure \ref{fig:score_dist}). We therefore create a simple maximum difference algorithm inspired by~\citet{salloum2017automated} to identify the best segmentation boundaries . We evaluate how likely each token is the introduction's start position by averaging the scores of tokens before and after it:
\begin{equation}
P_i = \frac{\sum_{n=1}^k{S_{i+n}}}{k} - \frac{\sum_{n=1}^k{S_{i-n}}}{k}
\end{equation}
where $P_i$ is the likelihood for a token to be the start position, $S_i$ is the \texttt{Is-intro} score assigned by our model, and $k$ is a chosen window size. The token $i$ that maximizes $P_i$ is chosen as the introduction start position. We select the end position of the introduction in a similar manner.

Additionally, we create a baseline using non-contextual word embeddings. We use GloVe vectors with 6B tokens and 100 dimensions provided by the Stanford NLP group~\cite{pennington2014glove}. Using the GloVe vectors for each token, we train a logistic classifier to predict the \texttt{Is-intro} probability for each token on the same training and test sets, and perform boundary detection using the method described above.

We use the pre-trained BERT model \texttt{bert-base-uncased} provided in the \texttt{Transformers} package~\cite{Wolf2019HuggingFacesTS}. We train the model for 300 epochs, using the \texttt{AdamW} optimizer with a target learning rate of $2e-5$. Linear warm-up and decay are used for learning rate adjustment. The cross entropy function is used as the loss function.

Automatic data augmentation techniques have been widely applied to alleviate the lack of labeled data in recent years~\cite{feng2021survey}. Here we experiment with two augmentation strategies that were found to consistently perform well~\cite{chen2021empirical}. The first is random word replacement based on TF--IDF scoring~\cite{xie2019unsupervised} (\textbf{tfidfwr}), and the second one is randomly applying token swap, deletion, or crop~\cite{wei2019eda} (\textbf{randomaug}). We experiment with different numbers of augmented samples and choose to generate 5 augmented samples for each original sample. The Python package \texttt{nlpaug}~\cite{ma2019nlpaug} is used to perform the augmentation.

\section{Results}\label{sec:results}

We begin by showing the overall alignment between each models' predictions and the ground truth in Figure \ref{fig:score_dist}. 
We first examine the base BERT model and the BERT models with data augmentation. 
We found that although all three models have a few misses as the base and word replacement models assign a few tokens with low \emph{Is-intro} scores within the true introduction, and the random augmentation model assigns high scores to a few tokens before the beginning of the true introduction, they correctly identify blocks of text that mostly align with the ground truth. 
Meanwhile, our token-based baseline predicts high \texttt{Is-intro} probability tokens throughout the text and is not able to recognize a block with consistent high scores, much less a segment that overlaps with the labeled introduction.

\begin{table*}[h]
\begin{center}
\footnotesize
\hspace*{-2cm}
\begin{tabular}{ccccccccccc}
\toprule
\multicolumn{11}{c}{Seen programs}\\
\hline
& \multicolumn{5}{c}{Intro start} & \multicolumn{5}{c}{Intro end} \\
\cmidrule(lr){2-6}\cmidrule(lr){7-11}
offset & 0 & 1 & 3 & 5 & 9 & 0 & 1 &3 &5 &9\\
\hline
base & 0.181 (0.148) & 0.219 (0.148) & 
0.324 (0.082) & 0.39 (0.108) & 0.486 (0.071) & 0.01 (0.041) & 0.086 (0.071) & \textbf{0.2 (0.071)} & 0.219 (0.041) & 0.286 (0.071)\\
tfidfwr  & 0.19 (0.108) & \textbf{0.248 (0.108)} & \textbf{0.324 (0.041)} & 0.41 (0.108) & \textbf{0.514 (0.071)} & \textbf{0.038 (0.041)} & \textbf{0.105 (0.148)} & 0.2 (0.123) & \textbf{0.267 (0.082)} & \textbf{0.343 (0.071)}\\
randaug & \textbf{0.2 (0.123)} & \textbf{0.248 (0.108)} & 0.305 (0.108) & \textbf{0.41 (0.082)} & 0.505 (0.179) & 0.01 (0.041) & 0.029 (0.071) & 0.086 (0.071) & 0.124 (0.082) & 0.219 (0.249)\\
glove & 0.029 & 0.057 & 0.114 & 0.2 & 0.257 & 0.0 & 0.0289 & 0.029 & 0.029 & 0.086 \\
\midrule
\multicolumn{11}{c}{Unseen programs}\\
\midrule
& \multicolumn{5}{c}{Intro start} & \multicolumn{5}{c}{Intro end} \\
\cmidrule(lr){2-6}\cmidrule(lr){7-11}
offset & 0 & 1 & 3 & 5 & 9 & 0 & 1 &3 &5 &9\\
\hline
base & 0.012 (0.051) & 0.024 (0.051) & 0.179 (0.54) & 0.214 (0.461) & 0.214 (0.461) & \textbf{0.012 (0.051)} & 0.012 (0.051) & 0.012 (0.051) & 0.024 (0.102) & 0.071 (0.154)\\
tfidfwr & 0.024 (0.102) & 0.024 (0.102) & 0.107 (0.235) & 0.131 (0.312) & 0.143 (0.32) & 0.0 (0) & 0.024 (0.102) & 0.071 (0.235) & 0.107 (0.235) & 0.202 (0.312)
 \\
randaug  & \textbf{0.095 (0.185)} &  \textbf{0.155 (0.205)} & \textbf{0.238 (0.205)} & \textbf{0.238 (0.205)} & \textbf{0.238 (0.205)} & 0.0 (0) & \textbf{0.036 (0.089)} & \textbf{0.119 (0.051)} & \textbf{0.179 (0)} & \textbf{0.345 (0.185)}
\\
glove & 0.0 & 0.0 & 0.0 & 0.0 & 0.0 & 0.0 & 0.0 & 0.0
& 0.0 & 0.036\\

\bottomrule
\end{tabular}
\caption{Accuracy versus offset on the test sets with seen and unseen programs, where the offset is how many tokens there are between the predicted position and the true position. The average results across 3 different runs are shown for the BERT models. 95\% confidence intervals are shown in parenthesis.}
\label{table:start_end_acc}
\end{center}
\end{table*}

With a general understanding of the models' behaviour, we formally evaluate our model's performance by two metrics inspired by the practice in~\citet{rajpurkar2016squad}. First, we examine the accuracy in predicting the segmentation boundaries. 
We consider the accuracy with regard to given offsets, i.e. how far away the predicted boundary is from the ground truth, in number of tokens. 
We consider offsets from 0 to 9; with the noisy and swift nature of podcasts, 9 spoken words happen in 2-3 seconds, and is a small margin compared to an entire podcast which can be longer than an hour.

The accuracy corresponding to varied offset is shown in Table \ref{table:start_end_acc}. 
We found that the BERT with word replacement in general has the best performance on the test set of seen programs, reaching an accuracy over 51\% at offset $= 9$ and 25\% over the GloVe baseline when predicting the start position. 
It also has an accuracy over 34\% at predicting the end position. 
Meanwhile, the BERT model with random augmentation performs best on the test set of unseen programs, potentially more resilient to unfamiliar data due to the variation introduced by augmentation. 
Notably, it also performs better in predicting the end position than predicting the start position, even though the end is harder to identify for other models as well as human readers (c.f. Section 2). 
Overall, our models are able to generalize into data that is structurally different from the training data, while the baseline generalizes poorly.

\begin{figure}[ht]
    \centering
    \begin{subfigure}{0.45\textwidth}
        \includegraphics[width=\textwidth]{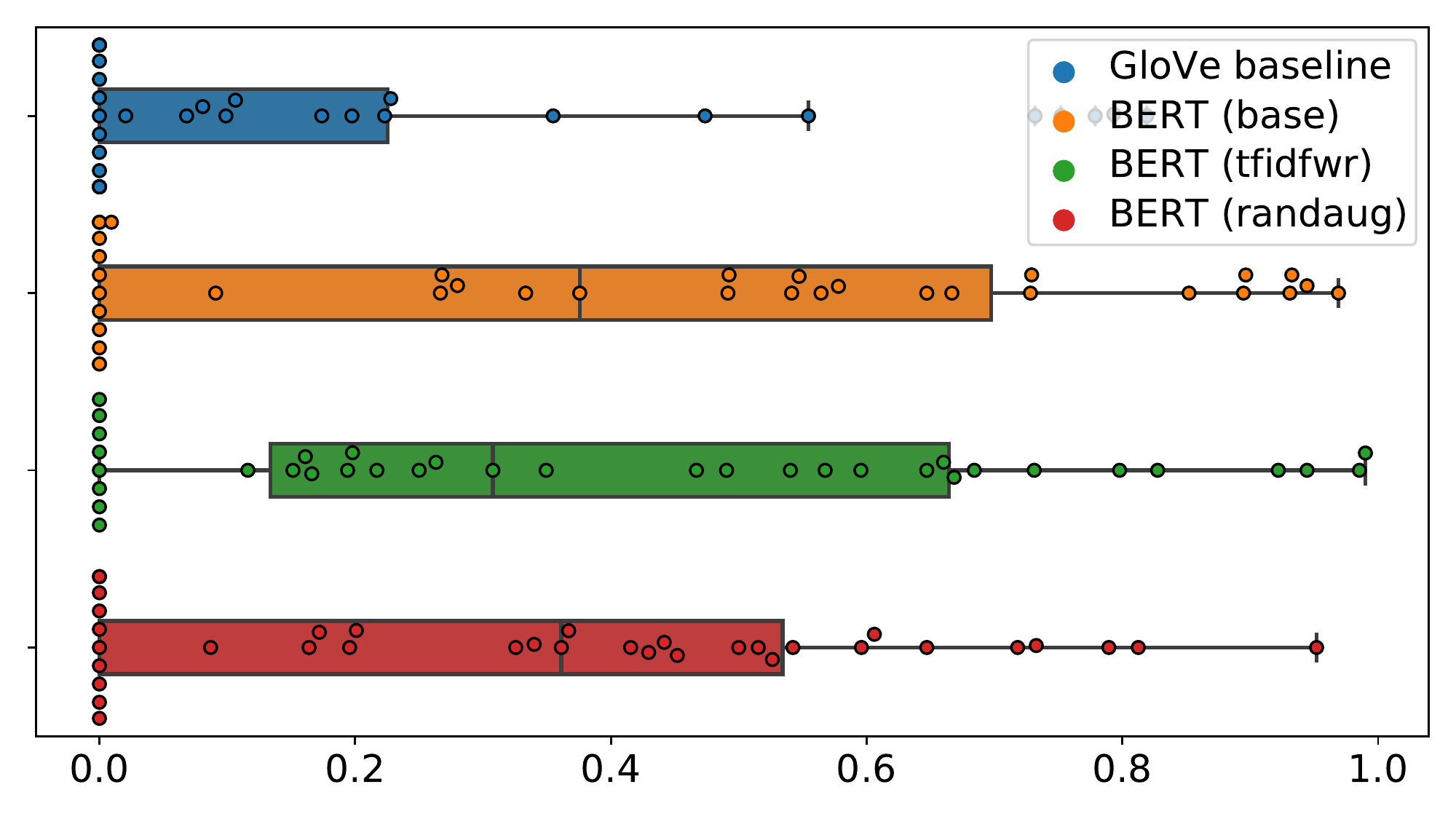}
        \vspace{-0.5cm}
        \caption{}
        \label{fig:overlap_swarm_seen}
    \end{subfigure}
    \begin{subfigure}{0.45\textwidth}
        \includegraphics[width=\textwidth]{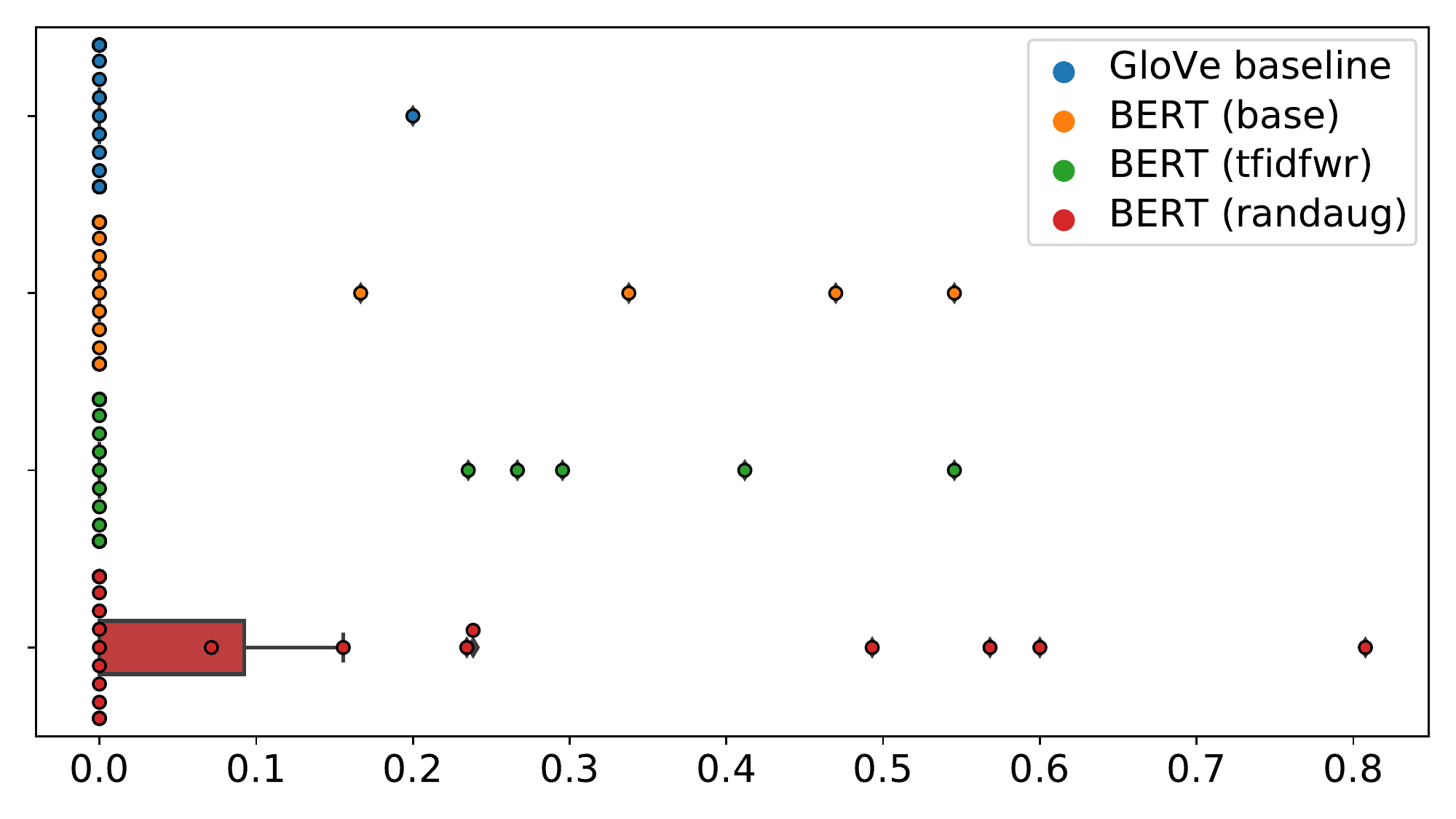}
         \vspace{-0.5cm}
        \caption{}
        \label{fig:overlap_swarm_unseen}
    \end{subfigure}
    \caption{Distribution of overlap scores. The x-axis shows the overlap score between 0 and 1. Each point is a document in the test sets. (a) shows the test data from previously seen programs, and (b) shows the test data from unseen programs.}
    \label{fig:overlap_swarm}
\end{figure}

We also evaluate the performance of our models using the overlap between the predicted introduction and true introduction on a token level. Because good predicted introductions need to not only contain the right tokens, but also be at the correct position, it is not suitable to compute the F1 score using a bag-of-tokens model. We compute the overlap score as:
\begin{equation}
S = \frac{\textrm{predicted intro} \cap \textrm{true intro} }{\textrm{predicted intro} \cup\textrm{true intro}}
\vspace{2ex}
\end{equation}

The overlap score distribution are summarized in Figure \ref{fig:overlap_swarm}. We found that although all BERT models work well on the test set of seen programs, the word replacement model has a significantly higher first quartile. However, on the test set of unseen programs, the random augmentation model fares better while all other models perform poorly, confirming our observation above.

\begin{figure}[ht]
    \centering
    \begin{subfigure}{0.45\textwidth}
\includegraphics[width=\textwidth]{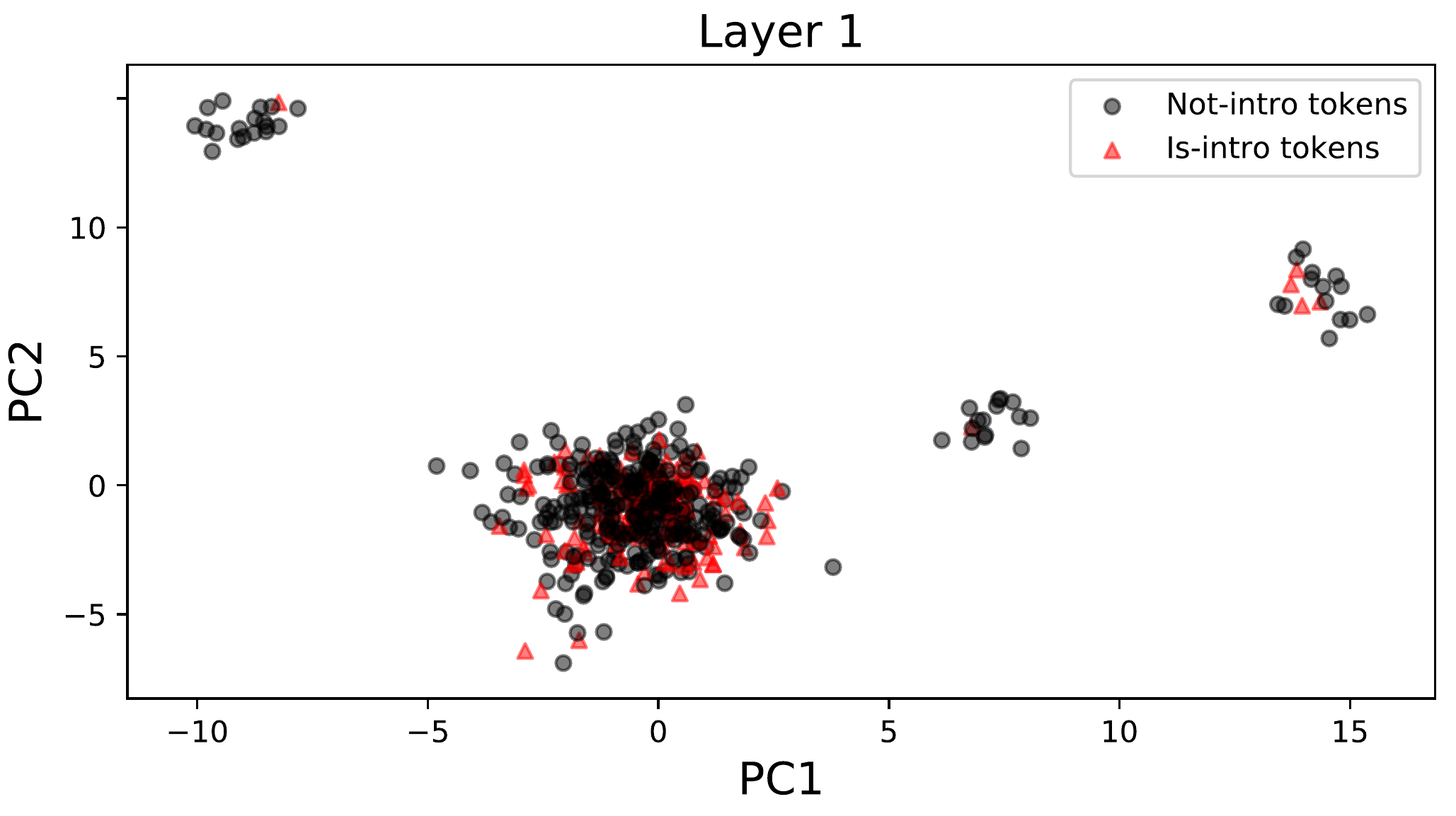}
    \vspace{-0.5cm}
        \caption{}
        \label{fig:bert_layer_1}
    \end{subfigure}
    \begin{subfigure}{0.465\textwidth}
        \includegraphics[width=\textwidth, ]{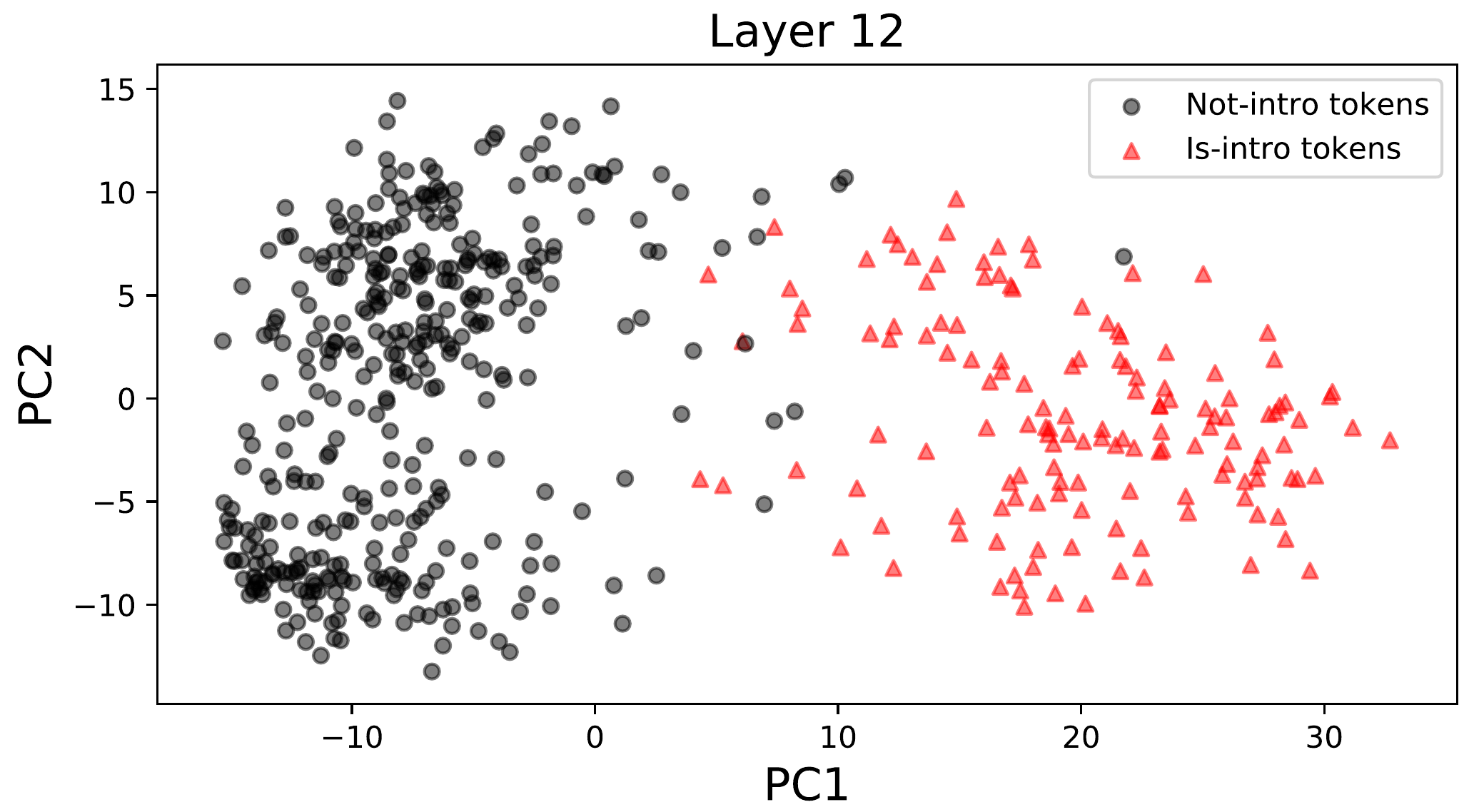}
        \caption{}
        \label{fig:bert_layer_12}
    \end{subfigure}
    \caption{Token clustering from the base BERT model's outputs. (a) output of layer 1. (b) output of layer 12. Red markers show the \texttt{Is-intro} tokens, and black show the \texttt{Not-intro} tokens.}
    \label{fig:bert_layers}
\end{figure}

We further analyze our models to better understand their learning behaviour. We extract the output of each hidden layer using the base BERT model as an example. Inspired by the method used in \citet{van2019does}, we perform a principal component analysis (PCA) on the output token vectors, which have 768 dimensions each. We then plot the tokens using the first two principal components. In this way, we expect to find clusters of tokens that the model considers to be closely related. Figure \ref{fig:bert_layers} shows the token clustering from output of layer 1 and layer 12. We found that in layer 1, words with similar syntactic functions are close to each other; for example, there is a cluster of verbs on the upper left corner and one of auxiliary verbs on the right. Meanwhile, there is no separation between the \texttt{Is-intro} or \texttt{Not-intro} tokens. However, at layer 12, two distinct clusters appear, containing the \texttt{Is-intro} and \texttt{Not-intro} tokens respectively. This is consistent with the findings in \citet{van2019does} and \citet{tenney2019bert}, where the lower and higher layers in BERT are found to contain different types of information. In the lower layers, such as layer 1, the model learns and maintains syntactic representation, while the higher layers focus on task-specific information. We speculate that such learning phases allow our models the versatility to learn structural information in addition to lexical and topic information.

\section{Discussion}\label{sec:discussion}

In this work, we proposed the task of identifying introductions in podcast episodes as a type of segment that can be used for listeners to sample content as well as for recommendation. We created a novel dataset of annotated ASR transcripts for our task, and developed Transformer-based models that agree well with human judgement in identifying the boundaries of introductions. Our models may be integrated into a pipeline that includes human validation of automatically discovered segments and uses them for downstream tasks.

The main challenge in our task is that the structures of the introductions are both not conventional, and greatly varied across different podcast genres. Although we did find that some programs had very regular introduction structure (e.g. {\it Song Exploder}), many had a looser, more conversational style, without a standardized structure. Additionally, transcription error is also a factor in our dataset, as is the fundamental difference between written and speech data. The speech data is generally less organized, less grammatically well-formed, and also error-prone, with restarts, disfluencies, over-talking, and the like.

We demonstrate that our model handles these challenges well. In particular, our model significantly out-performs the baseline on the test set of unseen programs, showing an ability to extend to unfamiliar data. This result also suggest that although the definition of an introduction is not crisp, there exists some embedded linguistic knowledge that human readers leverage to recognize structure, which the deep learning models are also able to learn.

Our dataset is relatively small in scale due to the limit in labeling resources, and only a subset was labeled by multiple annotators. The noise in ASR data poses additional challenges to the annotators, and not all annotators agree on the positions of introductions even with corrected transcripts. Despite these limitations, our dataset is one of the first labeled dataset for the structural segmentation of ASR transcripts. Compared to synthetic datasets constructed by concatenating different segments, our dataset is more internally coherent and challenging for machine learning models. We believe that our dataset will be a beneficial addition for future work in this domain, including better characterization of the narrative structure and stylistic variation of podcasts which would allow us to identify other stable segment types within spoken word programming.




\bibliographystyle{ACM-Reference-Format}
\bibliography{sample-base}


\begin{thebibliography}{30}


\ifx \showCODEN    \undefined \def \showCODEN     #1{\unskip}     \fi
\ifx \showDOI      \undefined \def \showDOI       #1{#1}\fi
\ifx \showISBNx    \undefined \def \showISBNx     #1{\unskip}     \fi
\ifx \showISBNxiii \undefined \def \showISBNxiii  #1{\unskip}     \fi
\ifx \showISSN     \undefined \def \showISSN      #1{\unskip}     \fi
\ifx \showLCCN     \undefined \def \showLCCN      #1{\unskip}     \fi
\ifx \shownote     \undefined \def \shownote      #1{#1}          \fi
\ifx \showarticletitle \undefined \def \showarticletitle #1{#1}   \fi
\ifx \showURL      \undefined \def \showURL       {\relax}        \fi
\providecommand\bibfield[2]{#2}
\providecommand\bibinfo[2]{#2}
\providecommand\natexlab[1]{#1}
\providecommand\showeprint[2][]{arXiv:#2}

\bibitem[\protect\citeauthoryear{Amat, Chandrashekar, Jebara, and
  Basilico}{Amat et~al\mbox{.}}{2018}]%
        {amat2018netflix}
\bibfield{author}{\bibinfo{person}{Fernando Amat}, \bibinfo{person}{Ashok
  Chandrashekar}, \bibinfo{person}{Tony Jebara}, {and} \bibinfo{person}{Justin
  Basilico}.} \bibinfo{year}{2018}\natexlab{}.
\newblock \showarticletitle{Artwork Personalization at Netflix}. In
  \bibinfo{booktitle}{\emph{Proceedings of the 12th ACM Conference on
  Recommender Systems}} (Vancouver, British Columbia, Canada)
  \emph{(\bibinfo{series}{RecSys ’18})}. \bibinfo{publisher}{Association for
  Computing Machinery}, \bibinfo{address}{New York, NY, USA},
  \bibinfo{pages}{487–488}.
\newblock
\showISBNx{9781450359016}
\urldef\tempurl%
\url{https://doi.org/10.1145/3240323.3241729}
\showDOI{\tempurl}


\bibitem[\protect\citeauthoryear{Amazon}{Amazon}{2018}]%
        {rekognition}
\bibfield{author}{\bibinfo{person}{Amazon}.} \bibinfo{year}{2018}\natexlab{}.
\newblock \bibinfo{title}{Amazon Rekognition}.
\newblock
\newblock
\urldef\tempurl%
\url{https://aws.amazon.com/rekognition/}
\showURL{%
\tempurl}


\bibitem[\protect\citeauthoryear{Badjatiya, Kurisinkel, Gupta, and
  Varma}{Badjatiya et~al\mbox{.}}{2018}]%
        {badjatiya2018attention}
\bibfield{author}{\bibinfo{person}{Pinkesh Badjatiya},
  \bibinfo{person}{Litton~J. Kurisinkel}, \bibinfo{person}{Manish Gupta}, {and}
  \bibinfo{person}{Vasudeva Varma}.} \bibinfo{year}{2018}\natexlab{}.
\newblock \showarticletitle{Attention-based Neural Text Segmentation}. In
  \bibinfo{booktitle}{\emph{European Conference on Information Retrieval}}.
  Springer, \bibinfo{pages}{180--193}.
\newblock


\bibitem[\protect\citeauthoryear{Bouchekif, Damnati, Est{\`{e}}ve, Charlet, and
  Camelin}{Bouchekif et~al\mbox{.}}{2015}]%
        {bouchekif2015diachronic}
\bibfield{author}{\bibinfo{person}{Abdessalam Bouchekif},
  \bibinfo{person}{G{\'{e}}raldine Damnati}, \bibinfo{person}{Yannick
  Est{\`{e}}ve}, \bibinfo{person}{Delphine Charlet}, {and}
  \bibinfo{person}{Nathalie Camelin}.} \bibinfo{year}{2015}\natexlab{}.
\newblock \showarticletitle{Diachronic Semantic Cohesion for Topic Segmentation
  of {TV} Broadcast News}. In \bibinfo{booktitle}{\emph{{INTERSPEECH} 2015,
  16th Annual Conference of the International Speech Communication Association,
  Dresden, Germany, September 6-10, 2015}}. \bibinfo{publisher}{{ISCA}},
  \bibinfo{pages}{2932--2936}.
\newblock
\urldef\tempurl%
\url{http://www.isca-speech.org/archive/interspeech\_2015/i15\_2932.html}
\showURL{%
\tempurl}


\bibitem[\protect\citeauthoryear{Carlson, Marcu, and Okurowsky}{Carlson
  et~al\mbox{.}}{2001}]%
        {carlson2001discourse}
\bibfield{author}{\bibinfo{person}{Lynn Carlson}, \bibinfo{person}{Daniel
  Marcu}, {and} \bibinfo{person}{Mary~Ellen Okurowsky}.}
  \bibinfo{year}{2001}\natexlab{}.
\newblock \showarticletitle{Building a Discourse-Tagged Corpus in the Framework
  of {R}hetorical {S}tructure {T}heory}. In
  \bibinfo{booktitle}{\emph{Proceedings of the Second {SIG}dial Workshop on
  Discourse and Dialogue}}.
\newblock
\urldef\tempurl%
\url{https://aclanthology.org/W01-1605}
\showURL{%
\tempurl}


\bibitem[\protect\citeauthoryear{Chen, Tam, Raffel, Bansal, and Yang}{Chen
  et~al\mbox{.}}{2021}]%
        {chen2021empirical}
\bibfield{author}{\bibinfo{person}{Jiaao Chen}, \bibinfo{person}{Derek Tam},
  \bibinfo{person}{Colin Raffel}, \bibinfo{person}{Mohit Bansal}, {and}
  \bibinfo{person}{Diyi Yang}.} \bibinfo{year}{2021}\natexlab{}.
\newblock \showarticletitle{An Empirical Survey of Data Augmentation for
  Limited Data Learning in {NLP}}.
\newblock \bibinfo{journal}{\emph{arXiv:2106.07499}} (\bibinfo{year}{2021}).
\newblock


\bibitem[\protect\citeauthoryear{Chifu and Fournier}{Chifu and
  Fournier}{2016}]%
        {chifu2016segchain}
\bibfield{author}{\bibinfo{person}{Adrian{-}Gabriel Chifu} {and}
  \bibinfo{person}{S{\'{e}}bastien Fournier}.} \bibinfo{year}{2016}\natexlab{}.
\newblock \showarticletitle{SegChain: Towards a Generic Automatic Video
  Segmentation Framework, based on Lexical Chains of Audio Transcriptions}. In
  \bibinfo{booktitle}{\emph{Proceedings of the 6th International Conference on
  Web Intelligence, Mining and Semantics, {WIMS} 2016, N{\^{\i}}mes, France,
  June 13-15, 2016}}, \bibfield{editor}{\bibinfo{person}{Rajendra Akerkar},
  \bibinfo{person}{Michel Planti{\'{e}}}, \bibinfo{person}{Sylvie Ranwez},
  \bibinfo{person}{S{\'{e}}bastien Harispe}, \bibinfo{person}{Anne Laurent},
  \bibinfo{person}{Patrice Bellot}, \bibinfo{person}{Jacky Montmain}, {and}
  \bibinfo{person}{Fran{\c{c}}ois Trousset}} (Eds.).
  \bibinfo{publisher}{{ACM}}, \bibinfo{pages}{21:1--21:8}.
\newblock
\urldef\tempurl%
\url{https://doi.org/10.1145/2912845.2912872}
\showDOI{\tempurl}


\bibitem[\protect\citeauthoryear{Cohan, Dernoncourt, Kim, Bui, Kim, Chang, and
  Goharian}{Cohan et~al\mbox{.}}{2018}]%
        {cohan2018discourse}
\bibfield{author}{\bibinfo{person}{Arman Cohan}, \bibinfo{person}{Franck
  Dernoncourt}, \bibinfo{person}{Doo~Soon Kim}, \bibinfo{person}{Trung Bui},
  \bibinfo{person}{Seokhwan Kim}, \bibinfo{person}{Walter Chang}, {and}
  \bibinfo{person}{Nazli Goharian}.} \bibinfo{year}{2018}\natexlab{}.
\newblock \showarticletitle{A Discourse-aware Attention Model for Abstractive
  Summarization of Long Documents}.
\newblock \bibinfo{journal}{\emph{arXiv:1804.05685}} (\bibinfo{year}{2018}).
\newblock


\bibitem[\protect\citeauthoryear{Covington, Adams, and Sargin}{Covington
  et~al\mbox{.}}{2016}]%
        {covington2016youtube}
\bibfield{author}{\bibinfo{person}{Paul Covington}, \bibinfo{person}{Jay
  Adams}, {and} \bibinfo{person}{Emre Sargin}.}
  \bibinfo{year}{2016}\natexlab{}.
\newblock \showarticletitle{Deep Neural Networks for YouTube Recommendations}.
  In \bibinfo{booktitle}{\emph{Proceedings of the 10th ACM Conference on
  Recommender Systems}} (Boston, Massachusetts, USA)
  \emph{(\bibinfo{series}{RecSys ’16})}. \bibinfo{publisher}{Association for
  Computing Machinery}, \bibinfo{address}{New York, NY, USA},
  \bibinfo{pages}{191–198}.
\newblock
\showISBNx{9781450340359}
\urldef\tempurl%
\url{https://doi.org/10.1145/2959100.2959190}
\showDOI{\tempurl}


\bibitem[\protect\citeauthoryear{Devlin, Chang, Lee, and Toutanova}{Devlin
  et~al\mbox{.}}{2018}]%
        {devlin2018bert}
\bibfield{author}{\bibinfo{person}{Jacob Devlin}, \bibinfo{person}{Ming{-}Wei
  Chang}, \bibinfo{person}{Kenton Lee}, {and} \bibinfo{person}{Kristina
  Toutanova}.} \bibinfo{year}{2018}\natexlab{}.
\newblock \showarticletitle{{BERT:} Pre-training of Deep Bidirectional
  Transformers for Language Understanding}.
\newblock  (\bibinfo{year}{2018}).
\newblock
\showeprint[arxiv]{1810.04805}


\bibitem[\protect\citeauthoryear{Feng, Gangal, Wei, Chandar, Vosoughi,
  Mitamura, and Hovy}{Feng et~al\mbox{.}}{2021}]%
        {feng2021survey}
\bibfield{author}{\bibinfo{person}{Steven~Y Feng}, \bibinfo{person}{Varun
  Gangal}, \bibinfo{person}{Jason Wei}, \bibinfo{person}{Sarath Chandar},
  \bibinfo{person}{Soroush Vosoughi}, \bibinfo{person}{Teruko Mitamura}, {and}
  \bibinfo{person}{Eduard Hovy}.} \bibinfo{year}{2021}\natexlab{}.
\newblock \showarticletitle{A Survey of Data Augmentation Approaches for
  {NLP}}.
\newblock \bibinfo{journal}{\emph{arXiv:2105.03075}} (\bibinfo{year}{2021}).
\newblock


\bibitem[\protect\citeauthoryear{Hovy}{Hovy}{1993}]%
        {hovy1993automated}
\bibfield{author}{\bibinfo{person}{Eduard~H. Hovy}.}
  \bibinfo{year}{1993}\natexlab{}.
\newblock \showarticletitle{Automated Discourse Generation Using Discourse
  Structure Relations}.
\newblock \bibinfo{journal}{\emph{Artificial intelligence}}
  \bibinfo{volume}{63}, \bibinfo{number}{1-2} (\bibinfo{year}{1993}),
  \bibinfo{pages}{341--385}.
\newblock


\bibitem[\protect\citeauthoryear{Hsueh and Moore}{Hsueh and Moore}{2007}]%
        {hsueh2010combining}
\bibfield{author}{\bibinfo{person}{Pei-Yun Hsueh} {and}
  \bibinfo{person}{Johanna~D. Moore}.} \bibinfo{year}{2007}\natexlab{}.
\newblock \showarticletitle{Combining Multiple Knowledge Sources for Dialogue
  Segmentation in Multimedia Archives}.
\newblock  (\bibinfo{date}{June} \bibinfo{year}{2007}),
  \bibinfo{pages}{1016--1023}.
\newblock
\urldef\tempurl%
\url{https://aclanthology.org/P07-1128}
\showURL{%
\tempurl}


\bibitem[\protect\citeauthoryear{Li, Sun, and Joty}{Li et~al\mbox{.}}{2018}]%
        {li2018segbot}
\bibfield{author}{\bibinfo{person}{Jing Li}, \bibinfo{person}{Aixin Sun}, {and}
  \bibinfo{person}{Shafiq~R. Joty}.} \bibinfo{year}{2018}\natexlab{}.
\newblock \showarticletitle{SegBot: A Generic Neural Text Segmentation Model
  with Pointer Network.}. In \bibinfo{booktitle}{\emph{Proceedings of the
  Twenty-Seventh International Joint Conference on Artificial Intelligence,
  {IJCAI-18}}}. \bibinfo{publisher}{International Joint Conferences on
  Artificial Intelligence Organization}, \bibinfo{pages}{4166--4172}.
\newblock
\urldef\tempurl%
\url{https://doi.org/10.24963/ijcai.2018/579}
\showDOI{\tempurl}


\bibitem[\protect\citeauthoryear{Ma}{Ma}{2019}]%
        {ma2019nlpaug}
\bibfield{author}{\bibinfo{person}{Edward Ma}.}
  \bibinfo{year}{2019}\natexlab{}.
\newblock \bibinfo{title}{NLP Augmentation}.
\newblock \bibinfo{howpublished}{https://github.com/makcedward/nlpaug}.
\newblock


\bibitem[\protect\citeauthoryear{Mochales and Moens}{Mochales and
  Moens}{2011}]%
        {mochales2011argumentation}
\bibfield{author}{\bibinfo{person}{Raquel Mochales} {and}
  \bibinfo{person}{Marie-Francine Moens}.} \bibinfo{year}{2011}\natexlab{}.
\newblock \showarticletitle{Argumentation Mining}.
\newblock \bibinfo{journal}{\emph{Artificial Intelligence and Law}}
  \bibinfo{volume}{19}, \bibinfo{number}{1} (\bibinfo{year}{2011}),
  \bibinfo{pages}{1--22}.
\newblock


\bibitem[\protect\citeauthoryear{Pak and Teh}{Pak and Teh}{2018}]%
        {pak2018text}
\bibfield{author}{\bibinfo{person}{Irina Pak} {and} \bibinfo{person}{Phoey~Lee
  Teh}.} \bibinfo{year}{2018}\natexlab{}.
\newblock \showarticletitle{Text Segmentation Techniques: A Critical Review}.
\newblock In \bibinfo{booktitle}{\emph{Innovative Computing, Optimization and
  Its Applications}}. Vol.~\bibinfo{volume}{741}.
  \bibinfo{publisher}{Springer}, \bibinfo{pages}{167--181}.
\newblock
\showISBNx{978-3-319-66983-0}
\urldef\tempurl%
\url{https://doi.org/10.1007/978-3-319-66984-7_10}
\showDOI{\tempurl}


\bibitem[\protect\citeauthoryear{Pennington, Socher, and Manning}{Pennington
  et~al\mbox{.}}{2014}]%
        {pennington2014glove}
\bibfield{author}{\bibinfo{person}{Jeffrey Pennington},
  \bibinfo{person}{Richard Socher}, {and} \bibinfo{person}{Christopher
  Manning}.} \bibinfo{year}{2014}\natexlab{}.
\newblock \showarticletitle{{G}lo{V}e: Global Vectors for Word Representation}.
  In \bibinfo{booktitle}{\emph{Proceedings of the 2014 Conference on Empirical
  Methods in Natural Language Processing ({EMNLP})}}.
  \bibinfo{publisher}{Association for Computational Linguistics},
  \bibinfo{address}{Doha, Qatar}, \bibinfo{pages}{1532--1543}.
\newblock
\urldef\tempurl%
\url{https://doi.org/10.3115/v1/D14-1162}
\showDOI{\tempurl}


\bibitem[\protect\citeauthoryear{Purver, K{\"o}rding, Griffiths, and
  Tenenbaum}{Purver et~al\mbox{.}}{2006}]%
        {purver2006unsupervised}
\bibfield{author}{\bibinfo{person}{Matthew Purver}, \bibinfo{person}{Konrad~P.
  K{\"o}rding}, \bibinfo{person}{Thomas~L. Griffiths}, {and}
  \bibinfo{person}{Joshua~B. Tenenbaum}.} \bibinfo{year}{2006}\natexlab{}.
\newblock \showarticletitle{Unsupervised Topic Modelling for Multi-Party Spoken
  Discourse}. In \bibinfo{booktitle}{\emph{Proceedings of the 21st
  International Conference on Computational Linguistics and 44th Annual Meeting
  of the Association for Computational Linguistics}}.
  \bibinfo{publisher}{Association for Computational Linguistics},
  \bibinfo{address}{Sydney, Australia}, \bibinfo{pages}{17--24}.
\newblock
\urldef\tempurl%
\url{https://doi.org/10.3115/1220175.1220178}
\showDOI{\tempurl}


\bibitem[\protect\citeauthoryear{Rajpurkar, Zhang, Lopyrev, and
  Liang}{Rajpurkar et~al\mbox{.}}{2016}]%
        {rajpurkar2016squad}
\bibfield{author}{\bibinfo{person}{Pranav Rajpurkar}, \bibinfo{person}{Jian
  Zhang}, \bibinfo{person}{Konstantin Lopyrev}, {and} \bibinfo{person}{Percy
  Liang}.} \bibinfo{year}{2016}\natexlab{}.
\newblock \showarticletitle{SQuAD: 100,000+ Questions for Machine Comprehension
  of Text}.
\newblock  (\bibinfo{year}{2016}).
\newblock
\showeprint[arxiv]{1606.05250}


\bibitem[\protect\citeauthoryear{Rotman, Porat, Ashour, and Barzelay}{Rotman
  et~al\mbox{.}}{2018}]%
        {rotman2018video}
\bibfield{author}{\bibinfo{person}{Daniel Rotman}, \bibinfo{person}{Dror
  Porat}, \bibinfo{person}{Gal Ashour}, {and} \bibinfo{person}{Udi Barzelay}.}
  \bibinfo{year}{2018}\natexlab{}.
\newblock \showarticletitle{Optimally Grouped Deep Features Using Normalized
  Cost for Video Scene Detection}. In \bibinfo{booktitle}{\emph{Proceedings of
  the 2018 ACM on International Conference on Multimedia Retrieval}} (Yokohama,
  Japan) \emph{(\bibinfo{series}{ICMR ’18})}. \bibinfo{publisher}{Association
  for Computing Machinery}, \bibinfo{address}{New York, NY, USA},
  \bibinfo{pages}{187–195}.
\newblock
\showISBNx{9781450350464}
\urldef\tempurl%
\url{https://doi.org/10.1145/3206025.3206055}
\showDOI{\tempurl}


\bibitem[\protect\citeauthoryear{Salloum, Finley, Edwards, Miller, and
  Suendermann{-}Oeft}{Salloum et~al\mbox{.}}{2017}]%
        {salloum2017automated}
\bibfield{author}{\bibinfo{person}{Wael Salloum}, \bibinfo{person}{Greg
  Finley}, \bibinfo{person}{Erik Edwards}, \bibinfo{person}{Mark Miller}, {and}
  \bibinfo{person}{David Suendermann{-}Oeft}.} \bibinfo{year}{2017}\natexlab{}.
\newblock \showarticletitle{Automated Preamble Detection in Dictated Medical
  Reports}. In \bibinfo{booktitle}{\emph{BioNLP 2017, Vancouver, Canada, August
  4, 2017}}, \bibfield{editor}{\bibinfo{person}{Kevin~Bretonnel Cohen},
  \bibinfo{person}{Dina Demner{-}Fushman}, \bibinfo{person}{Sophia Ananiadou},
  {and} \bibinfo{person}{Junichi Tsujii}} (Eds.).
  \bibinfo{publisher}{Association for Computational Linguistics},
  \bibinfo{pages}{287--295}.
\newblock
\urldef\tempurl%
\url{https://doi.org/10.18653/v1/W17-2336}
\showDOI{\tempurl}


\bibitem[\protect\citeauthoryear{Tenney, Das, and Pavlick}{Tenney
  et~al\mbox{.}}{2019}]%
        {tenney2019bert}
\bibfield{author}{\bibinfo{person}{Ian Tenney}, \bibinfo{person}{Dipanjan Das},
  {and} \bibinfo{person}{Ellie Pavlick}.} \bibinfo{year}{2019}\natexlab{}.
\newblock \showarticletitle{{BERT} Rediscovers the Classical {NLP} Pipeline}.
\newblock \bibinfo{journal}{\emph{arXiv:1905.05950}} (\bibinfo{year}{2019}).
\newblock


\bibitem[\protect\citeauthoryear{van Aken, Winter, L{\"o}ser, and Gers}{van
  Aken et~al\mbox{.}}{2019}]%
        {van2019does}
\bibfield{author}{\bibinfo{person}{Betty van Aken}, \bibinfo{person}{Benjamin
  Winter}, \bibinfo{person}{Alexander L{\"o}ser}, {and}
  \bibinfo{person}{Felix~A. Gers}.} \bibinfo{year}{2019}\natexlab{}.
\newblock \showarticletitle{How Does {BERT} Answer Questions?: A Layer-Wise
  Analysis of Transformer Representations}. In
  \bibinfo{booktitle}{\emph{Proceedings of the 28th ACM International
  Conference on Information and Knowledge Management}}. ACM,
  \bibinfo{pages}{1823--1832}.
\newblock


\bibitem[\protect\citeauthoryear{Wang, Li, and Yang}{Wang
  et~al\mbox{.}}{2018}]%
        {wangetal2018toward}
\bibfield{author}{\bibinfo{person}{Yizhong Wang}, \bibinfo{person}{Sujian Li},
  {and} \bibinfo{person}{Jingfeng Yang}.} \bibinfo{year}{2018}\natexlab{}.
\newblock \showarticletitle{Toward Fast and Accurate Neural Discourse
  Segmentation}. In \bibinfo{booktitle}{\emph{Proceedings of the 2018
  Conference on Empirical Methods in Natural Language Processing}}.
  \bibinfo{publisher}{Association for Computational Linguistics},
  \bibinfo{address}{Brussels, Belgium}, \bibinfo{pages}{962–967}.
\newblock
\urldef\tempurl%
\url{https://doi.org/10.18653/v1/D18-1116}
\showDOI{\tempurl}


\bibitem[\protect\citeauthoryear{Webber, Prasad, Lee, and Joshi}{Webber
  et~al\mbox{.}}{2019}]%
        {webber2019penn}
\bibfield{author}{\bibinfo{person}{Bonnie Webber}, \bibinfo{person}{Rashmi
  Prasad}, \bibinfo{person}{Alan Lee}, {and} \bibinfo{person}{Aravind Joshi}.}
  \bibinfo{year}{2019}\natexlab{}.
\newblock \showarticletitle{The {P}enn {D}iscourse {T}reebank 3.0 {A}nnotation
  {M}anual}.
\newblock \bibinfo{journal}{\emph{Philadelphia, University of Pennsylvania}}
  (\bibinfo{year}{2019}).
\newblock
\urldef\tempurl%
\url{https://doi.org/11272.1/AB2/SUU9CB}
\showDOI{\tempurl}


\bibitem[\protect\citeauthoryear{Wei and Zou}{Wei and Zou}{2019}]%
        {wei2019eda}
\bibfield{author}{\bibinfo{person}{Jason~W. Wei} {and} \bibinfo{person}{Kai
  Zou}.} \bibinfo{year}{2019}\natexlab{}.
\newblock \showarticletitle{{EDA}: Easy Data Augmentation Techniques for
  Boosting Performance on Text Classification Tasks}.
\newblock \bibinfo{journal}{\emph{arXiv:1901.11196}} (\bibinfo{year}{2019}).
\newblock


\bibitem[\protect\citeauthoryear{Wolf, Debut, Sanh, Chaumond, Delangue, Moi,
  Cistac, Rault, Louf, Funtowicz, and Brew}{Wolf et~al\mbox{.}}{2019}]%
        {Wolf2019HuggingFacesTS}
\bibfield{author}{\bibinfo{person}{Thomas Wolf}, \bibinfo{person}{Lysandre
  Debut}, \bibinfo{person}{Victor Sanh}, \bibinfo{person}{Julien Chaumond},
  \bibinfo{person}{Clement Delangue}, \bibinfo{person}{Anthony Moi},
  \bibinfo{person}{Pierric Cistac}, \bibinfo{person}{Tim Rault},
  \bibinfo{person}{R{\'{e}}mi Louf}, \bibinfo{person}{Morgan Funtowicz}, {and}
  \bibinfo{person}{Jamie Brew}.} \bibinfo{year}{2019}\natexlab{}.
\newblock \showarticletitle{HuggingFace's Transformers: State-of-the-art
  Natural Language Processing}.
\newblock  (\bibinfo{year}{2019}).
\newblock
\showeprint[arxiv]{1910.03771}


\bibitem[\protect\citeauthoryear{Xie, Dai, Hovy, Luong, and Le}{Xie
  et~al\mbox{.}}{2019}]%
        {xie2019unsupervised}
\bibfield{author}{\bibinfo{person}{Qizhe Xie}, \bibinfo{person}{Zihang Dai},
  \bibinfo{person}{Eduard Hovy}, \bibinfo{person}{Minh-Thang Luong}, {and}
  \bibinfo{person}{Quoc~V Le}.} \bibinfo{year}{2019}\natexlab{}.
\newblock \showarticletitle{Unsupervised Data Augmentation for Consistency
  Training}.
\newblock \bibinfo{journal}{\emph{arXiv:1904.12848}} (\bibinfo{year}{2019}).
\newblock


\bibitem[\protect\citeauthoryear{Xu, Gan, Cheng, and Liu}{Xu
  et~al\mbox{.}}{2019}]%
        {xu2019discourse}
\bibfield{author}{\bibinfo{person}{Jiacheng Xu}, \bibinfo{person}{Zhe Gan},
  \bibinfo{person}{Yu Cheng}, {and} \bibinfo{person}{Jingjing Liu}.}
  \bibinfo{year}{2019}\natexlab{}.
\newblock \showarticletitle{Discourse-Aware Neural Extractive Model for Text
  Summarization}.
\newblock \bibinfo{journal}{\emph{arXiv:1910.14142}} (\bibinfo{year}{2019}).
\newblock


\end{thebibliography}










\end{document}